\RequirePackage[svgnames,table]{xcolor}

\documentclass[11pt,letterpaper,logo]{yalearxiv}

\usepackage{graphicx}

\usepackage{xcolor}

\definecolor{GTTechGold}{HTML}{B3A369}      
\definecolor{GTBuzzGold}{HTML}{EAAA00}      
\definecolor{GTNavy}{HTML}{003057}          

\definecolor{GTLightGrey}{HTML}{F2F2F2}     
\definecolor{GTDarkGrey}{HTML}{545454}      
\definecolor{GTBlack}{HTML}{000000}         

\usepackage{float,epstopdf}
\usepackage{bbm}

\usepackage{microtype}

\usepackage{natbib}
\setcitestyle{square}

\usepackage{subcaption}
\usepackage{booktabs}

\usepackage{amsmath}
\usepackage{amssymb}
\usepackage{mathtools}
\usepackage{amsthm}
\usepackage{dsfont}
\usepackage{multicol}
\usepackage{makecell}
\usepackage{multirow} 
\usepackage{amsfonts} 
\usepackage{mathrsfs}
\usepackage[amssymb, thickqspace]{SIunits}
\usepackage{enumitem}
\usepackage{pgfplotstable}
\usepackage{lipsum}		

\usepackage{microtype}
\usepackage{graphicx}
\usepackage{booktabs} 
\usepackage[table]{xcolor}

\usepackage[normalem]{ulem} 

\usepackage{cases}
\usepackage{wrapfig}

\usepackage{url}

\usepackage{thmtools}
\usepackage{thm-restate}
\usepackage{tabu}

\definecolor{huskypurple}{HTML}{4B2E83}

\hypersetup{
    colorlinks=true,
    linkcolor=GTNavy,
    citecolor=GTNavy,
    urlcolor=GTNavy
}
\usepackage{titletoc}

\usepackage{listings}
\lstdefinestyle{promptstyle}{
  basicstyle=\ttfamily\footnotesize,
  breaklines=true,
  breakautoindent=false,
  breakindent=0pt,
  postbreak=\mbox{\textcolor{gray}{$\hookrightarrow$}\space},
  columns=fullflexible,
  keepspaces=true,
  frame=single,
  framesep=5pt,
  xleftmargin=6pt,
  xrightmargin=6pt,
  aboveskip=8pt,
  belowskip=8pt,
  showstringspaces=false,
}

\usepackage{booktabs}
\usepackage{longtable}
\usepackage{arydshln}   
\makeatletter
\def\munderbar#1{\underline{\sbox\tw@{$#1$}\dp\tw@\z@\box\tw@}}
\makeatother

\AddToHook{cmd/appendix/before}{%
  \setcounter{axiom}{0}%
}

\newcommand{\be}{\begin{equation}}
\newcommand{\ee}{\end{equation}}
\newcommand{\bea}{\begin{equation*}\begin{aligned}}
\newcommand{\eea}{\end{aligned}\end{equation*}}

\newtcolorbox{simpleElegantQuote}{
    colback=AliceBlue!50!White,
    colframe=RoyalBlue!75!Black,
    boxrule=0.5pt,
    arc=2mm,
    boxsep=4pt,
    left=10pt, right=10pt,
    top=8pt, bottom=8pt,
    fontupper=\itshape,
}

\title{THRIVE: \\ Therapeutic Humanoid Robot In Virtual Environment}

\runningtitle{THRIVE: Therapeutic Humanoid Robot In Virtual Environment}


\usepackage{fontawesome5}   

\definecolor{yaleblue}{RGB}{0,58,112}

\author{
  Jin Xu\textsuperscript{1,2,$\dagger$}, 
  Yu-Ping Chen\textsuperscript{3}, 
  Ayanna Howard\textsuperscript{2} \\
  \textsuperscript{1}Georgia Institute of Technology, 
  \textsuperscript{2}The Ohio State University, 
  \textsuperscript{3}Georgia State University \\
}

\hypersetup{colorlinks=true, linkcolor=blue!50!black, citecolor=blue!50!black,
            urlcolor=blue!50!black}

\begin{document}

\begin{abstract}
\vspace{-1mm}
{\centering\section*{Abstract}}
This paper presents THRIVE (\textbf{T}herapeutic \textbf{H}umanoid \textbf{R}obot \textbf{I}n \textbf{V}irtual \textbf{E}nvironment), an at-home rehabilitation platform that integrates a suite of virtual-reality upper-body rehabilitation games, a real-time camera-based motion-tracking system, and a socially interactive robot therapist. The system is designed for therapy and intervention in children with upper-limb motor impairments, which can be improved through consistent, task-specific practice. THRIVE features a set of newly designed, engaging games that target functional reaching, grasping, and object-manipulation movements through customizable popping, hitting, catching, and grabbing tasks, while the camera-based tracking system captures the child's kinematic performance during play. A robot therapist—deployable either as a physical robotic coach or as a remote-presence virtual agent—delivers adaptive, dynamic feedback to motivate the child and guide their movements toward therapeutic goals. THRIVE decouples the therapeutic games from the robot embodiment, extending the platform to support various embodiments and different robots within one modular system. This robot-agnostic design makes THRIVE affordable, scalable, and readily adaptable for sustained use in the home, offering a practical pathway to more consistent and engaging upper-limb therapy for children with motor function impairments.
\end{abstract}

\maketitle
\footnotetext[1]{$\dagger$ J. Xu is currently with the Georgia Institute of Technology, Atlanta, GA 30332 (email: jxu81@gatech.edu). Most of the development work of this platform was completed when he was a postdoctoral scholar at the Ohio State University. This research supported by the Field Initiated Program of the National Institute on Disability, Independent Living, and Rehabilitation Research (NIDILRR) under award number 90IFST0009.}

\section{Introduction}
\label{sec:intro}
Cerebral palsy (CP) is one of the most common neurological disorders in childhood, affecting approximately 1 in 345 children in the United States and an estimated 17 million individuals worldwide~\citep{durkin2016prevalence, graham2016erratum}. Children with CP often have difficulty controlling their muscles, resulting in impaired balance and coordination, as well as involuntary movements. The majority of children diagnosed with CP experience problems with upper-limb function, including difficulty with reaching and grasping~\citep{lloyd2024upper, makki2014prevalence, de2016kinematic}.

Interventions targeting arm function in children with CP typically focus on repetitive, goal-directed tasks performed on a regular basis. Studies have shown that consistent, task-specific practice can significantly improve motor function and functional independence (e.g., neuroplasticity; constraint-induced movement therapy)~\citep{novak2013systematic, chen2014effectiveness}. However, many children with CP and their caregivers struggle with adherence to these programs due to factors such as low motivation, boredom, time constraints, and limited feedback or support. As such, there is a need for a home-based intervention system that is engaging, motivating, and accessible, and easily integrated into home environments and daily routines.

To address these challenges, we developed the \textbf{T}herapeutic \textbf{H}umanoid \textbf{R}obot \textbf{I}n \textbf{V}irtual \textbf{E}nvironment (THRIVE) system. The THRIVE system features:
\begin{enumerate}
    \item a set of newly designed upper-body rehabilitation games targeting fundamental movements such as reaching, grasping, and object manipulation;
    \item an upper-body tracking system that captures and analyzes the child's kinematic performance in real time;
    \item an affordable, interactive robot therapist that provides interactive guidance and feedback.
\end{enumerate}

In this system, children engage in interactive exercise games while the tracking system analyzes their movement patterns and performance. The robot therapist acts as a motivational coach, guiding children through tasks, providing encouragement, and adapting the difficulty level based on individual performance. This integrated approach aims to improve engagement, increase adherence to therapy, and ultimately enhance upper-limb motor outcomes in children with CP. The THRIVE system is designed to be scalable and accessible, enabling more children with CP to benefit from consistent, high-quality therapy in home settings.


\section{Related Work}
\label{sec:related_work}

Conventional physical and occupational therapy for children with neurological disorders such as cerebral palsy (CP) relies on repetitive, therapist-guided exercises. This approach is often effective, but it is also labor-intensive, costly, and difficult to sustain given limited clinical access and children's low tolerance for repetitive tasks \citep{lopes2018games}. To address engagement and compliance, researchers introduced serious games as a therapeutic complement, embedding rehabilitation exercises into game mechanics to increase motivation without displacing conventional therapy \citep{lopes2018games, ahn2023scoping}. Building on this, virtual reality (VR) systems extended serious games with richer, more immersive environments and, in some cases, automated assessment. Super Pop VR\texttrademark{} \citep{garcia2013super}, for example, paired a VR bubble-popping game with an automated Fugl-Meyer-based scoring method to track upper-extremity motor function, validated to within 5\% error against clinician assessment. A systematic review and meta-analysis of randomized controlled trials found VR interventions to be an effective tool for improving motor and functional outcomes in children with CP, though effect sizes and adherence to motor-learning principles varied across studies \citep{chen2018effectiveness, demers2021integration}. Despite these advances, most serious game and VR systems for CP remain screen- or sensor-based (e.g., webcam, EyeToy, low-cost motion controllers) and lack any physically embodied or socially interactive component alongside the exercise itself.

In parallel, robots have been introduced into rehabilitation both as physical assistive devices and as socially interactive coaches. A systematic review of robotic therapy for upper-extremity function in children with CP found generally positive but heterogeneous evidence across a small number of studies \citep{chen2016effects}. Within this space, humanoid and socially assistive robots (SARs), including NAO, ZORA, MAKRO, and Ursus, have been used primarily as motivators for joint-mobility or therapeutic exercises rather than as sources of physical assistance \citep{jeglinsky2024rehabilitation}. A pilot study using the ZORA robot in children with severe physical disabilities demonstrated positive support for therapeutic and educational activities by boosting engagement and motivation \citep{van2017robot}. Low-cost humanoid platforms such as KineTron, built from a commercial robotics kit, have similarly demonstrated feasibility for motivating repetitive motor training in small samples of children with CP \citep{kozyavkin2014humanoid}. Most closely related to the present work is \citet{chen2018effect}, which paired the Super Pop VR game with a humanoid robot (DARwIn-OP) that delivered real-time verbal feedback during gameplay, and found that children with and without CP adjusted their reaching kinematics in response to the robot's feedback. A follow-up study found that social interaction, rather than robot appearance, was the primary driver of engagement gains when a different robot platform was substituted into the same protocol \citep{lee2017does}.


\section{Methodology}
\label{sec:method}

\begin{figure}[t]
  \centering
  \includegraphics[width=\columnwidth]{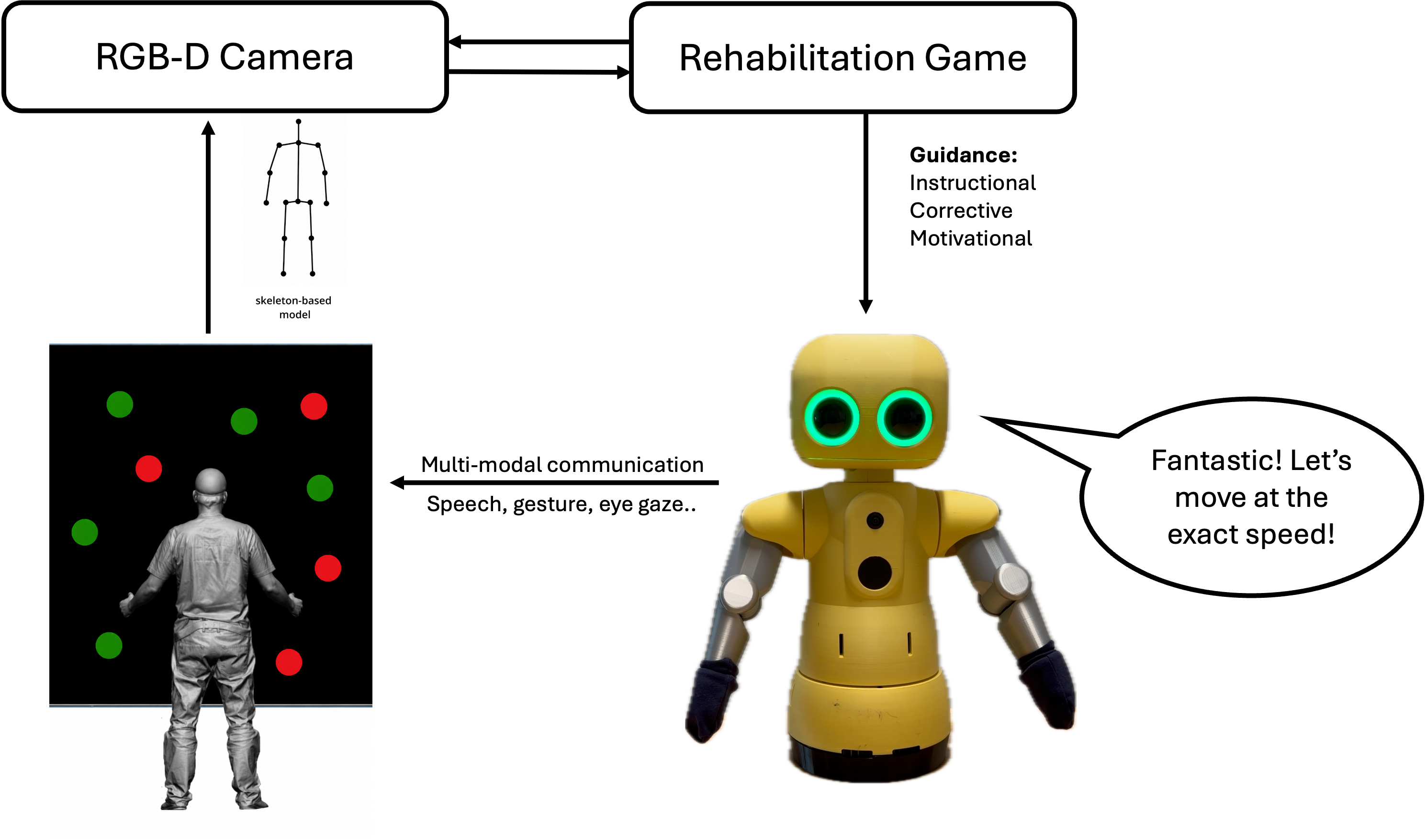}
  \caption{System Architecture}
  \label{fig:system}
\end{figure}

The THRIVE system consists of two major components: rehabilitation games and the THRIVE robot, which plays the role of an AI therapist and can be embodied as a physical robot, a remote-presence robot, or a virtual agent. Figure~\ref{fig:system} shows the overall system architecture of the THRIVE system.

\subsection{Games}

The THRIVE system comes with a suite of four newly developed games based on the original SuperPop games \citep{garcia2013super}. When playing these games, the child is immersed in a virtual world containing various virtual objects. These virtual objects are fully customizable, including their color, shape (bubble, square, or cartoon character), and speed of movement. With these customizable settings, the research team can adapt the placement of virtual objects to create appropriately challenging reaching tasks for each child.

We designed four different games: a popping game, a hitting game, a catching game, and a grabbing game. All four games require children to actively use their impacted arm to interact with virtual objects within a set time frame. For children with cerebral palsy (CP), these tasks are designed to exercise the affected arm through overhead, outward, and across-the-midline movement patterns, individually or in combination. Children with CP tend to spontaneously avoid these types of movements, yet such activities are important functional movements for daily life and are therefore central to their intervention and recovery. Accordingly, the games are designed to improve movement speed, reaching accuracy, and precision, with the overall goal of improving the child's long-term quality of life through improved function of the affected arm.

\textbf{Popping game.} The popping game is a remake of the original SuperPop game \citep{garcia2013super}, built with a newer game engine and updated tracking technology. It was designed to replace the original SuperPop game for clinical use, as the Microsoft Kinect has since been discontinued. Note that the games described below are variants of the popping game, each targeting a different type of reaching movement. In the popping game, we focus on activities that require children to use both arms simultaneously to reach for and hit one or more virtual targets that appear dynamically on screen at set intervals. Both green and red bubbles appear randomly on the screen, and children are instructed to use their arms to ``pop'' the green bubbles while avoiding the red ones within a specific time limit. When the time expires, the bubbles disappear.

\begin{figure}[h]
    \centering
    \includegraphics[width=0.6\linewidth]{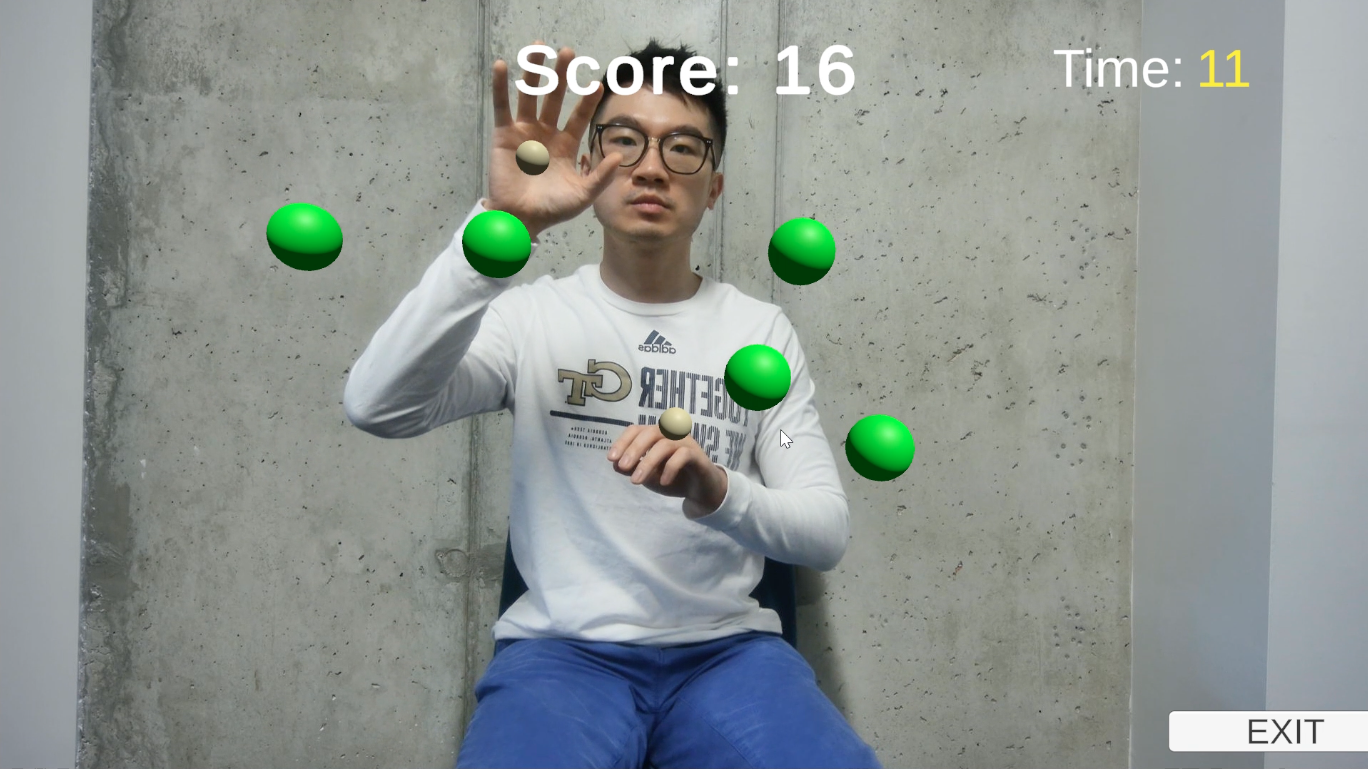}
    \caption{Popping game. A demonstration of using both arms simultaneously to pop green bubbles as they appear dynamically on screen.}
    \label{fig:popping_game}
\end{figure}

\textbf{Hitting game.} Whereas the popping game focuses on reaching activities using both arms, the hitting game focuses on completing a sequence of consecutive reaching movements. In this game, three to five bubbles appear in a row, and children are instructed to use their arms to strike through all the bubbles in sequence.

\begin{figure}[h]
    \centering
    \includegraphics[width=0.6\linewidth]{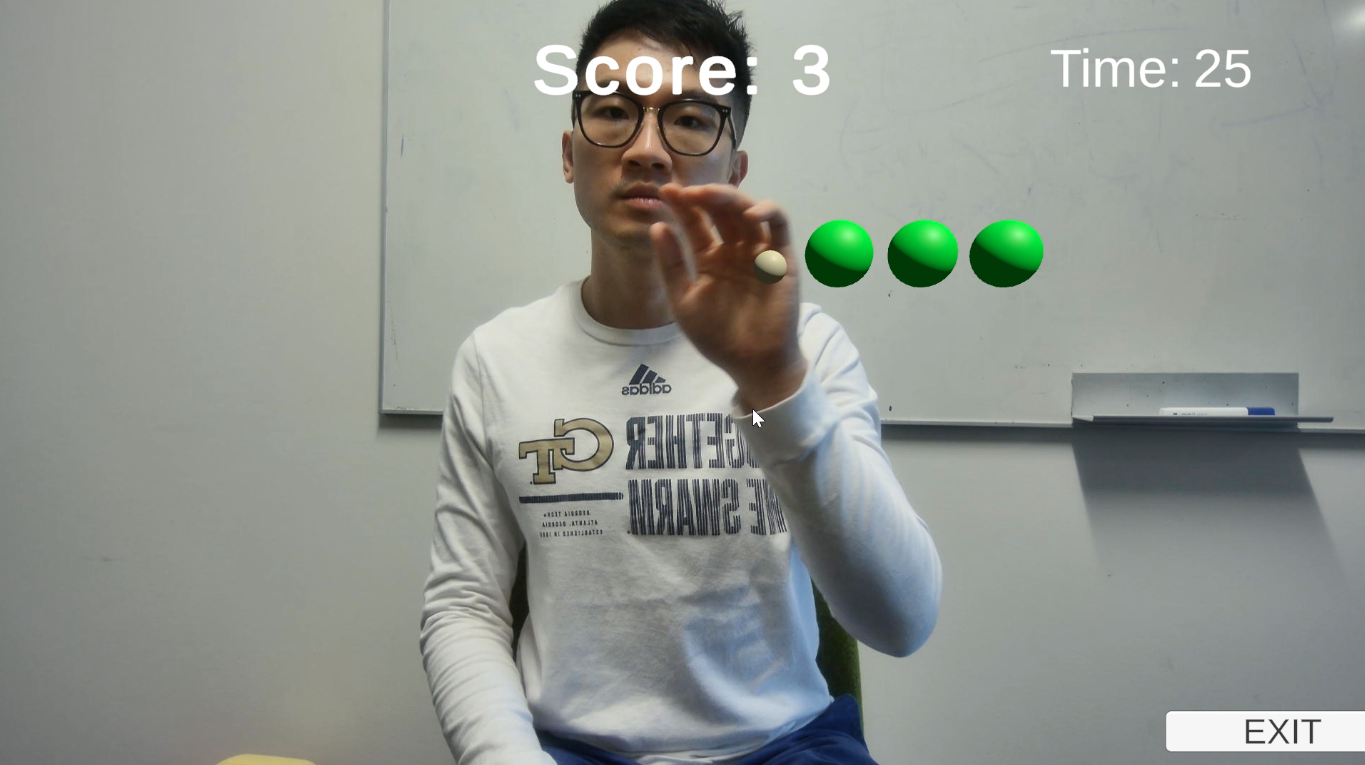}
    \caption{Hitting game. A demonstration of striking through a row of three to five bubbles in sequence, targeting consecutive reaching movements.}
    \label{fig:hitting_game}
\end{figure}

\textbf{Catching game.} Unlike the popping game, the catching game focuses on reaching for a single virtual object that appears randomly and moves dynamically across the screen. In this game, a bubble moves from one side of the screen to the other, and children are instructed to pop the moving bubble. Because the bubble moves along both horizontal and vertical axes, this game encourages the development of hand-eye coordination.

\begin{figure}[h]
    \centering
    \includegraphics[width=0.6\linewidth]{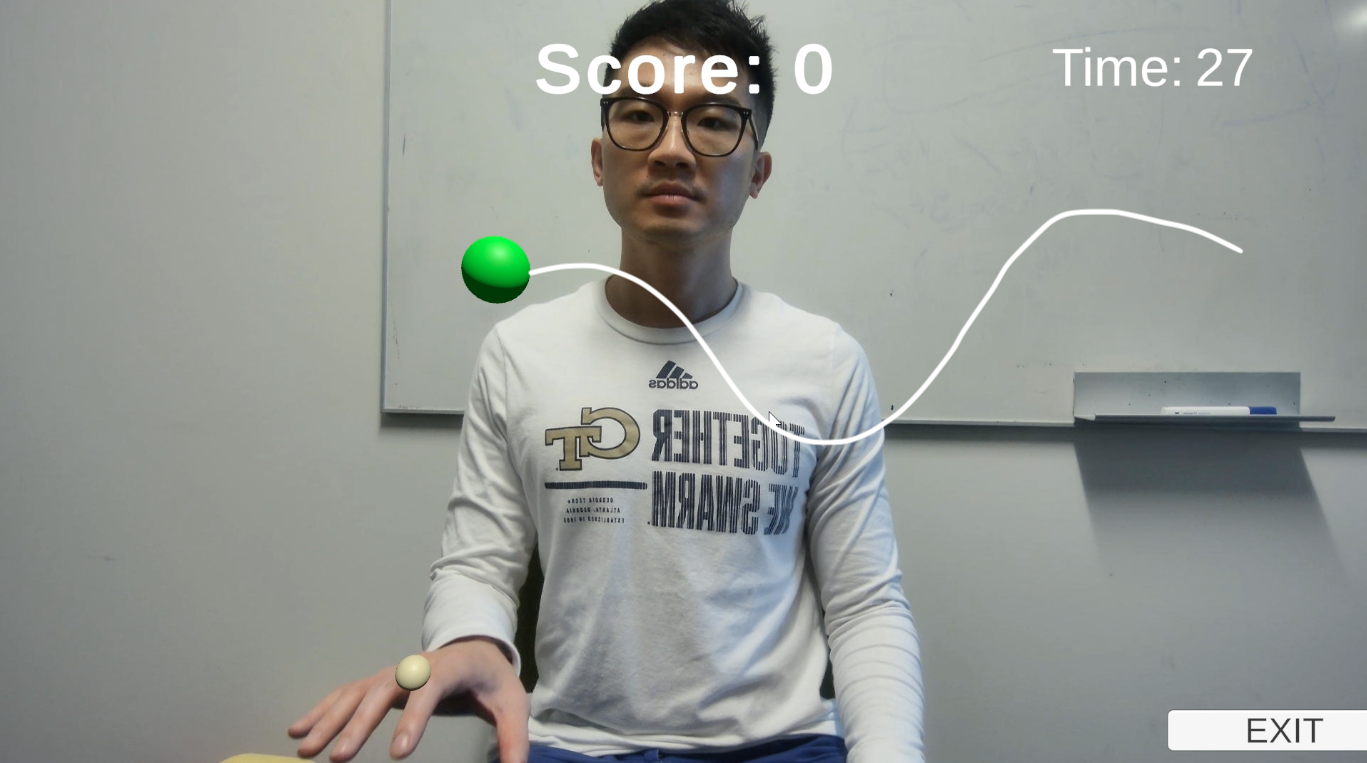}
    \caption{Catching game. An example trajectory illustrating the bubble's dynamic movement across the screen along both horizontal and vertical axes.}
    \label{fig:catching_game}
\end{figure}

\textbf{Grabbing game.} In the grabbing game, a bubble appears on one side of the screen and a designated box appears on the other side. Children are instructed to first ``touch'' the bubble, then grab it and drop it into the designated box. This task involves performing a virtual grasp-and-release action, similar to daily activities such as picking up an apple from one basket and placing it into another.

\begin{figure}[h]
    \centering
    \includegraphics[width=0.6\linewidth]{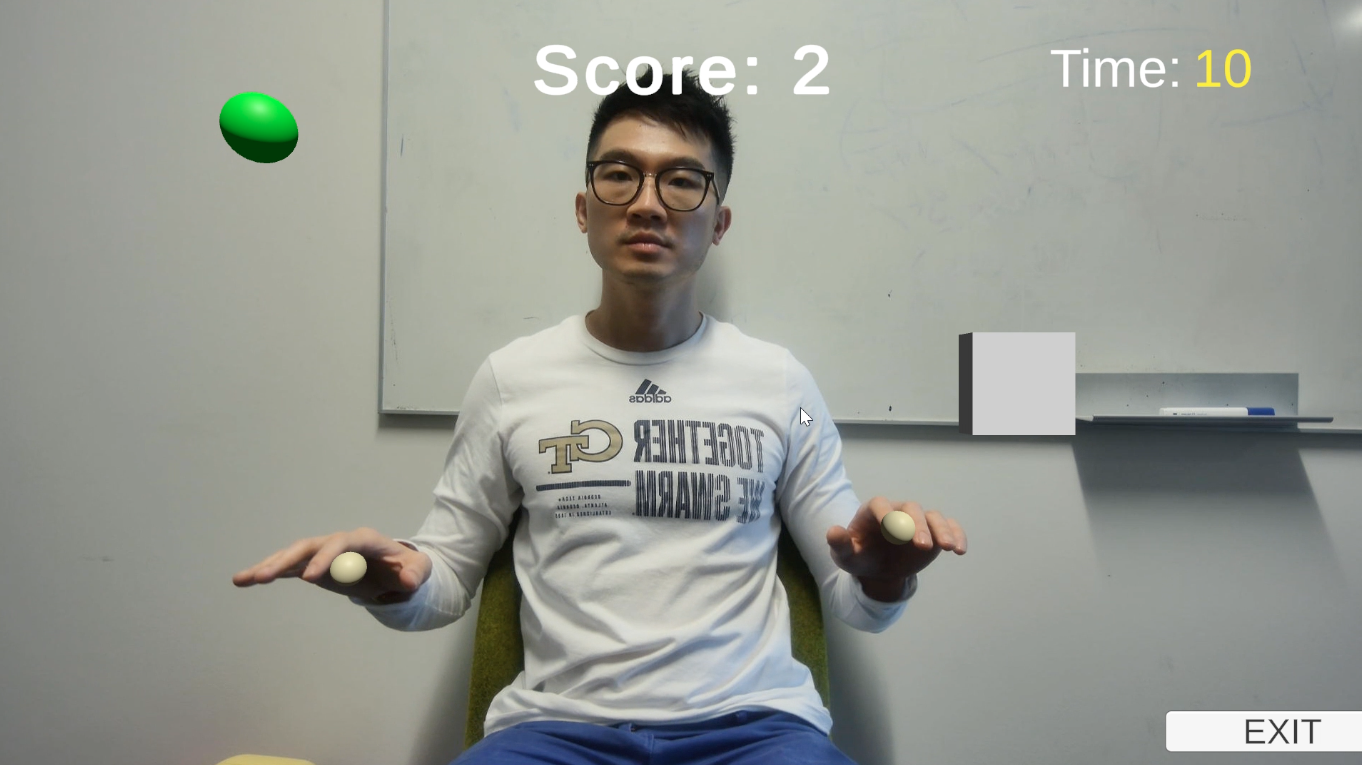}
    \caption{Grabbing game. A demonstration of touching, grabbing, and transporting a bubble from one side of the screen to a designated box on the other side, simulating a grasp-and-release task.}
    \label{fig:grabbing_game}
\end{figure}

\subsection{Camera Tracking System}

We originally used the \cite{microsoft_azure_kinect_dk} with the Body Tracking SDK to track users' body movements. After Microsoft announced the discontinuation of the Azure Kinect in 2023, the \cite{orbbec_femto_bolt} was adopted as a replacement camera, as it is based on similar technology and shares nearly identical specifications and performance.

\subsection{Robot Therapist}

The THRIVE system was designed to be accessible and affordable for all children with therapeutic needs. As such, we provide a variety of options for children to interact with the robot therapist.

\textbf{THRIVE Robot Therapist.} This robot was designed in partnership with the startup company \cite{hello_robotics}. The robot was 3D printed with swappable servo actuators for movement. It stands approximately 50 centimeters in height and weighs approximately 270 grams. The robot features a 7-degree-of-freedom (7-DOF) articulated body. Its motors are controlled using a Raspberry Pi 5, with external speakers and microphones enabling verbal communication. Additionally, as a social companion, the robot's eyes were designed with LED lights so that it can convey non-verbal social cues during interaction. The robot is programmed to exhibit a number of socially interactive behaviors, such as speaking and gesturing.

\begin{figure}[h]
    \centering
    \begin{subfigure}[b]{0.51\linewidth}
        \centering
        \includegraphics[width=\linewidth]{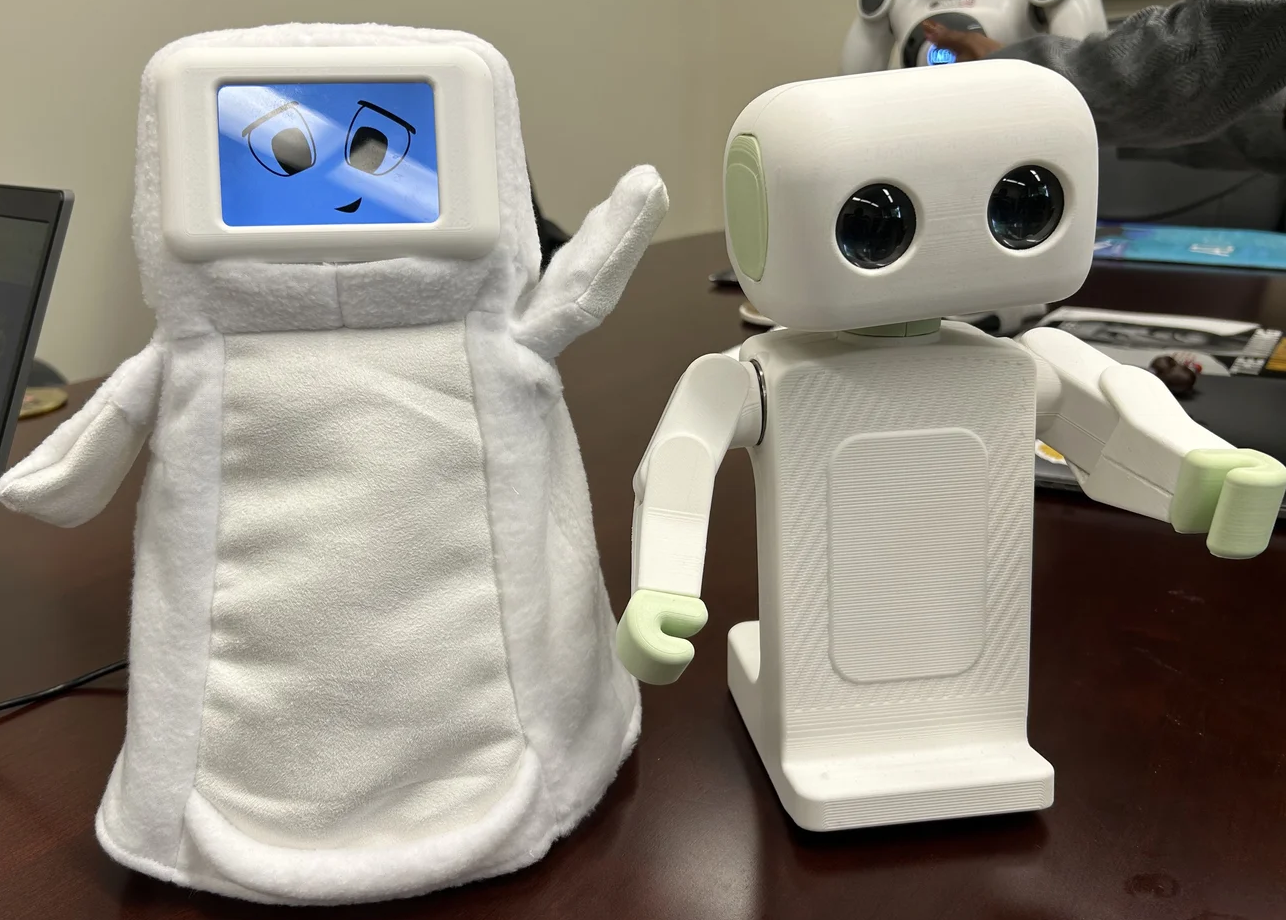}
        \caption{Early prototype}
        \label{fig:robot_prototype_early}
    \end{subfigure}
    \hspace{0.01\linewidth}
    \begin{subfigure}[b]{0.42\linewidth}
        \centering
        \includegraphics[width=\linewidth]{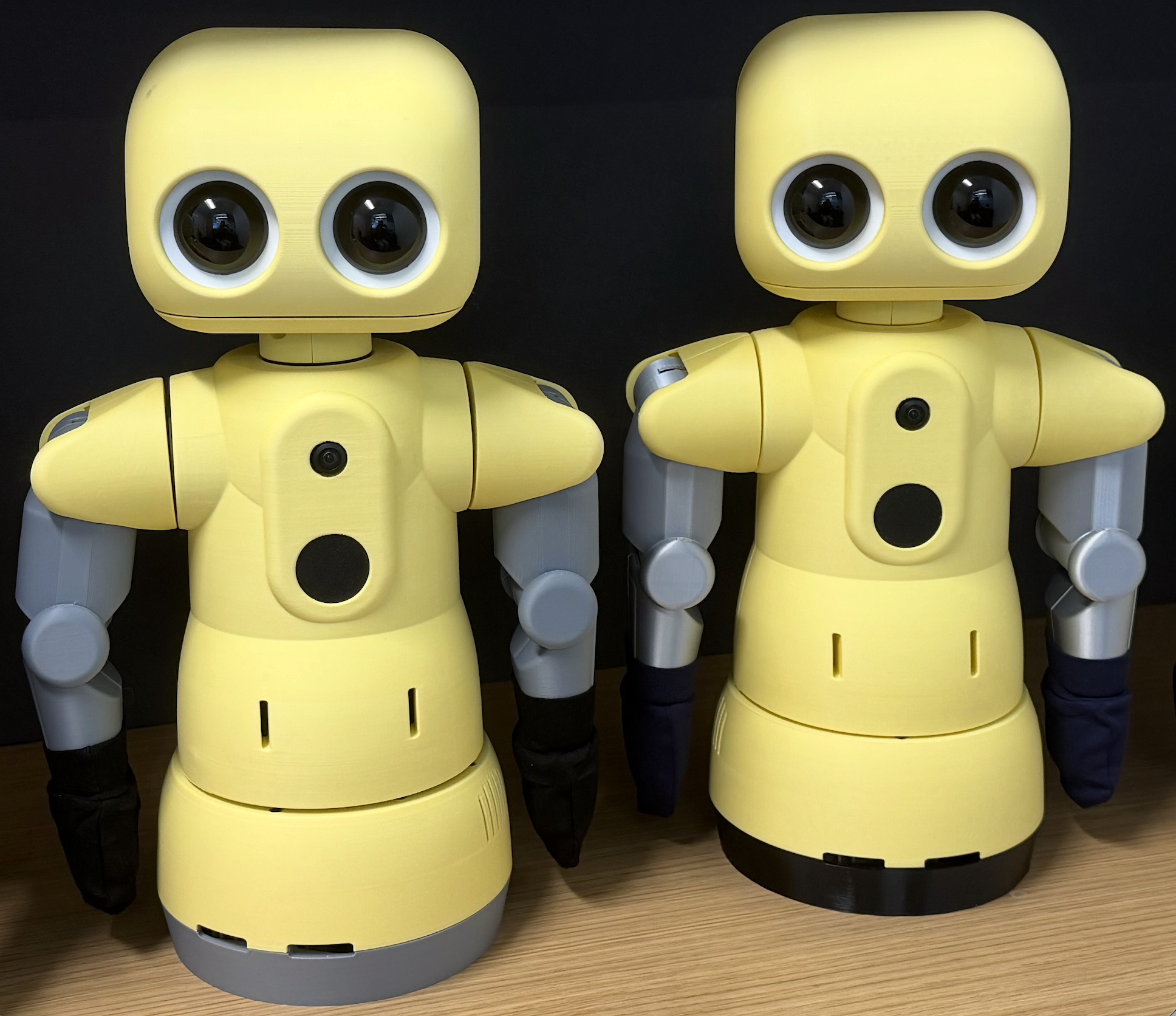}
        \caption{Final production robot}
        \label{fig:robot_prototype_final}
    \end{subfigure}
    \caption{From early prototype to production version of the THRIVE robot}
    \label{fig:robot_prototype}
\end{figure}

\textbf{Remote-Presence THRIVE Therapist.} To make the system more affordable for children who do not have access to a physical robot, we used pre-filmed videos of the physical robot together with a state-machine-based approach \citep{xu2018investigating} to enable remote presence. The robot performs the same behaviors during the interaction; the only difference is the mode of presence (physical vs. remote).

\begin{figure}[h]
    \centering
    \includegraphics[width=0.8\linewidth]{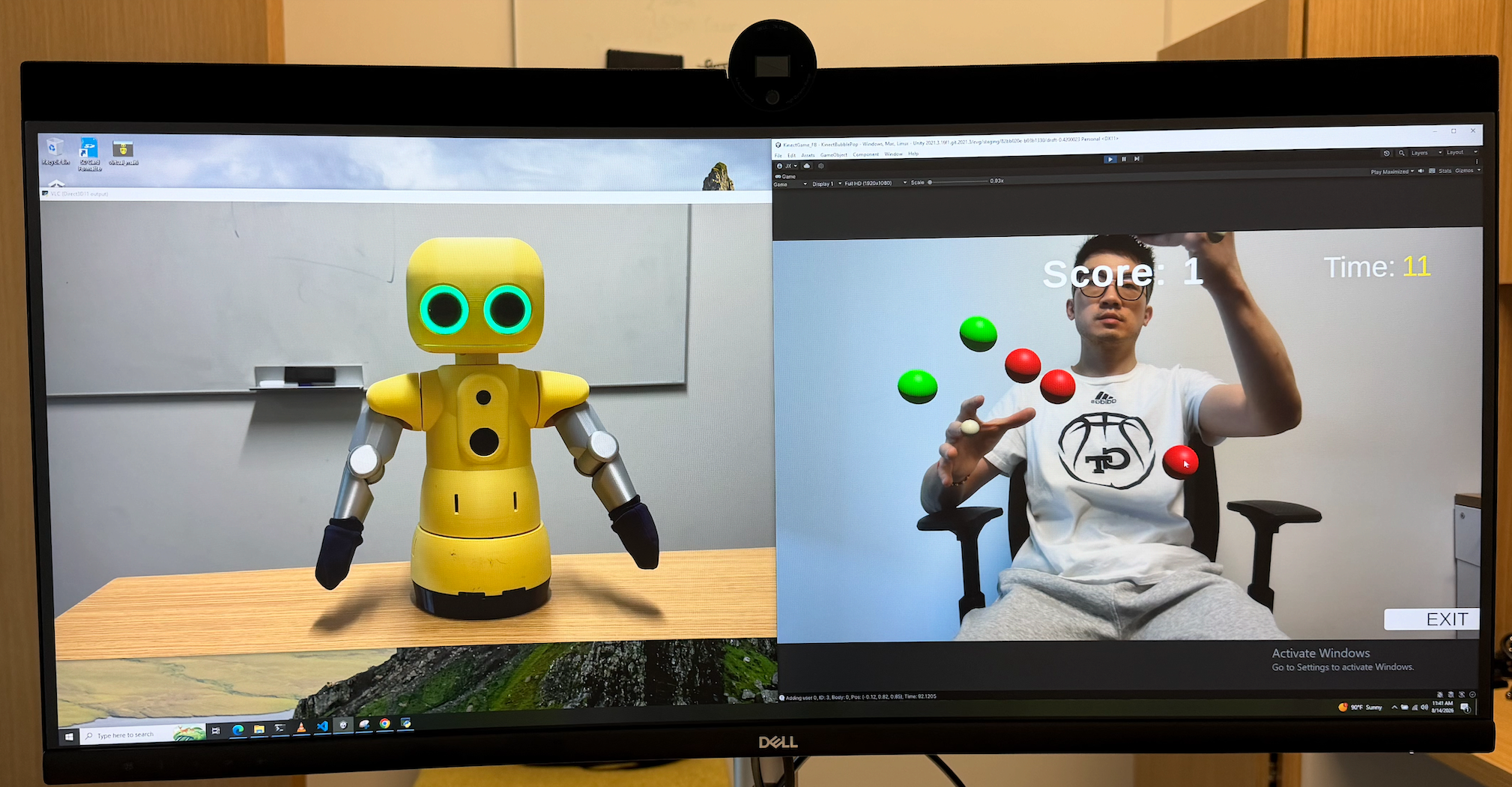}
    \caption{Remote Presence of THRIVE Robot}
    \label{fig:remove_agent}
\end{figure}

\textbf{THRIVE Virtual Agent.} In addition to the physical and remote-presence embodiments, we developed a virtual agent version of the THRIVE Robot to further expand accessibility for children who do not have access to a physical device or reliable video playback. The virtual agent was modeled directly from the CAD design of the physical robot, preserving its visual appearance, proportions, and articulated body structure. This allows the virtual agent to replicate the same verbal and non-verbal behaviors as the physical robot, including speech, gesturing, and LED-based eye cues, within a fully digital environment. The virtual agent can be rendered directly within the game environment, allowing it to be deployed on standard computers without requiring any additional hardware.

\begin{figure}[h]
    \centering
    \includegraphics[width=0.6\linewidth]{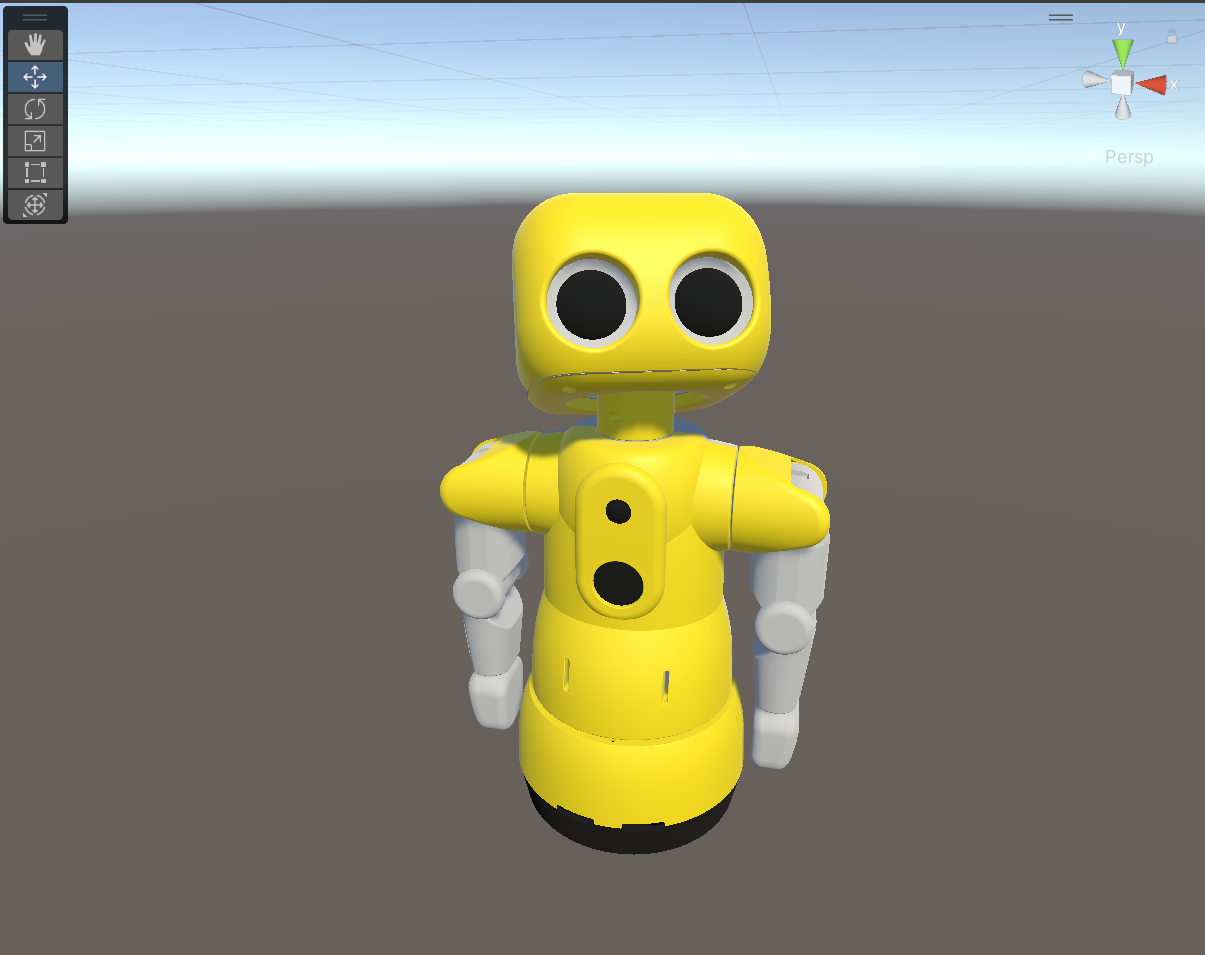}
    \caption{THRIVE virtual agent, modeled from the CAD design of the physical robot.}
    \label{fig:virtual_agent}
\end{figure}

\subsection{Robot Behaviors}

\textbf{Verbal communication.} We used the same script from the SuperPop-NAO platform \citep{lee2017does}, as it has been shown to be effective for rehabilitation in previous studies \citep{xu2018robot, de2019effect}. We then prompted GPT to generate many variations of these utterances while preserving the same meaning. These variations were passed to a text-to-speech system to produce the final verbal output delivered to the children. As a therapist, the robot primarily communicates with participants during three key phases of the interaction.

At the \textbf{beginning} of each game, the robot introduces itself and provides initial instructions using the following language: \textit{``Hello. My name is Thrive, and I will be playing SuperPop with you today. I will ask you to complete a series of tasks, and I would love it if you could follow my instructions. When you are ready, please raise both of your hands as high as you can.''}

\textbf{During} the game, the robot provides real-time guidance on the participant's movements. To encourage the participant to move faster, the robot says, \textit{``Keep up the good work. Move a little faster.''} To encourage the participant to slow down, the robot says, \textit{``Fantastic! Let us move at the exact same speed.''}

At the \textbf{end} of each game session, the robot provides positive feedback to the participant: \textit{``Fantastic! Wow! Good game. Let us play another one.''}

In the SuperPop-NAO study \citep{xu2018robot, de2019effect}, we used exactly the same language for the introduction and instructional feedback in every turn. However, feedback from that study indicated that children grew bored when the robot repeated the exact same wording across sessions. To address this, in the THRIVE platform, we prompted a large language model (GPT-4) \citep{openai2023gpt4} to generate variations of utterances for each of the three key phases, as shown in Tables~\ref{app:extra}, with the goal of increasing variation and naturalness in the robot's responses. These utterances are passed to a text-to-speech model for real-time verbal interaction.


\textbf{Non-verbal communication.} When the robot is about to speak to a child, its eyes change from the idle color (green) to the talking color (blue) to indicate that it is initiating communication. This behavior is particularly important for children, as it draws their attention in preparation for the next movement. Additionally, when talking, the robot moves its arms up and down slightly, emulating the natural hand gestures a human might use while giving a speech.

\begin{figure}[h]
    \centering
    \begin{subfigure}[b]{0.45\linewidth}
        \centering
        \includegraphics[width=\linewidth]{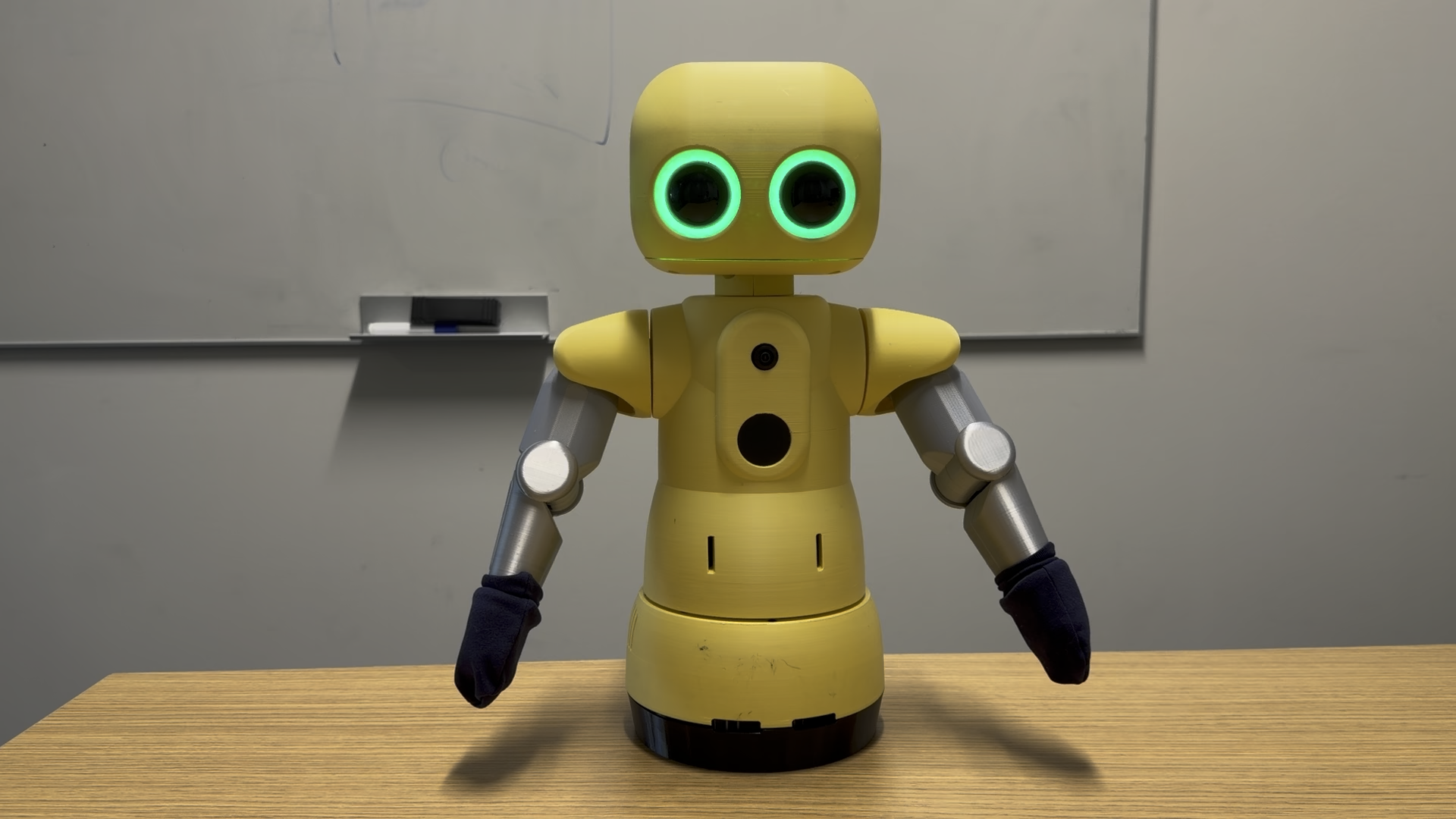}
        \caption{Standby mode.}
        \label{fig:nonverbal_standby}
    \end{subfigure}
    \hspace{0.01\linewidth}
    \begin{subfigure}[b]{0.45\linewidth}
        \centering
        \includegraphics[width=\linewidth]{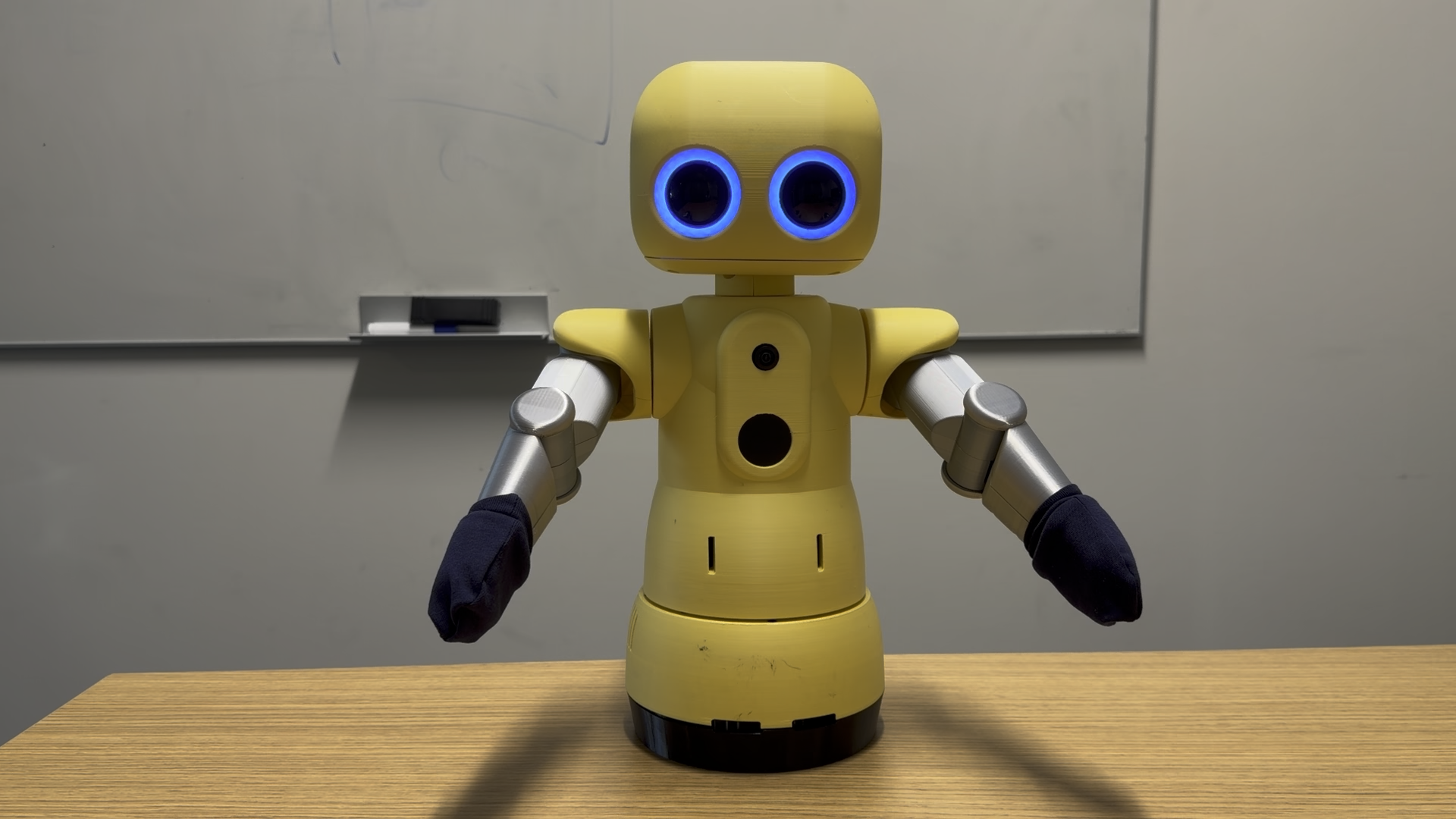}
        \caption{Talking mode.}
        \label{fig:nonverbal_talking}
    \end{subfigure}
    \caption{Non-verbal communication cues of the THRIVE robot}
    \label{fig:nonverbal_cues}
\end{figure}

\begin{figure}[h]
    \centering
    \includegraphics[width=1\linewidth]{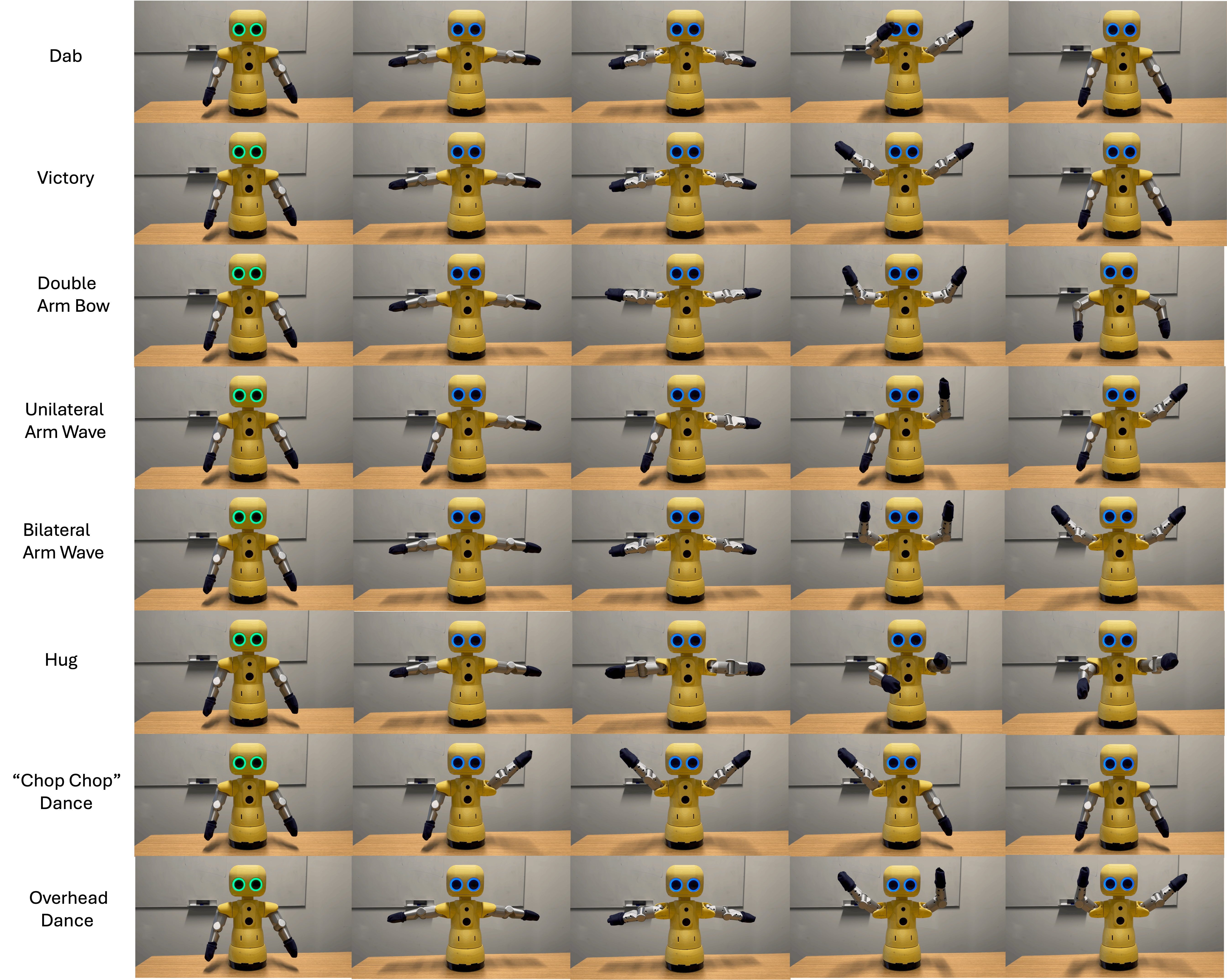}
    \caption{Celebratory movements to encourage children to play}
    \label{fig:dance}
\end{figure}

\textbf{Rehabilitation activities.} We programmed additional encouraging movements commonly used in physical therapy training programs as a way to motivate children to play another game. These celebratory movements (as shown in Figure \ref{fig:dance}) are played at the end of each game.

\section{Conclusion}
\label{sec:conclusion}

In this paper, we presented THRIVE, an embodiment-agnostic, at-home rehabilitation platform for children with cerebral palsy that pairs a suite of upper-body rehabilitation intervention games with a real-time camera-based tracking system and a socially interactive robot therapist. THRIVE decouples its virtual-reality serious-game layer from any single robot embodiment, allowing a physical robot, a virtual on-screen agent, or a remote-presence robot to serve interchangeably as the therapeutic coach, and prioritizes affordability and scalability for sustained home use rather than lab-based feasibility testing alone. By combining customizable, motor-learning-informed games, objective kinematic tracking, and an affordable, socially expressive robot capable of adaptive verbal and non-verbal feedback, THRIVE offers a low-cost pathway to consistent, engaging, and clinically grounded upper-limb therapy outside the clinic. Future work will evaluate THRIVE with children with CP across both physical and remote-presence robot configurations to assess engagement, adherence, and upper-limb motor outcomes, and to determine whether remote presence can serve as an effective, lower-cost substitute for in-home robot hardware at scale.


\section*{Acknowledgments}

This research supported by the Field Initiated Program of the National Institute on Disability, Independent Living, and Rehabilitation Research (NIDILRR) under award number 90IFST0009.

\bibliography{main}

@article{durkin2016prevalence,
  title={Prevalence of cerebral palsy among 8-year-old children in 2010 and preliminary evidence of trends in its relationship to low birthweight},
  author={Durkin, Maureen S and Benedict, Ruth E and Christensen, Deborah and Dubois, Lindsay A and Fitzgerald, Robert T and Kirby, Russell S and Maenner, Matthew J and Van Naarden Braun, Kim and Wingate, Martha S and Yeargin-Allsopp, Marshalyn},
  journal={Paediatric and perinatal epidemiology},
  volume={30},
  number={5},
  pages={496--510},
  year={2016},
  publisher={Wiley Online Library}
}

@article{graham2016erratum,
  title={Erratum: cerebral palsy},
  author={Graham, H Kerr and Rosenbaum, Peter and Paneth, Nigel and Dan, Bernard and Lin, Jean-Pierre and Damiano, Diane L and Becher, Jules G and Gaebler-Spira, Deborah and Colver, Allan and Reddihough, Dinah S and others},
  journal={Nature reviews disease primers},
  volume={2},
  number={1},
  pages={16005},
  year={2016},
  publisher={Nature Publishing Group}
}

@article{lloyd2024upper,
  title={Upper limb function in children with cerebral palsy: a structured approach to assessment and management},
  author={Lloyd, Jo and Yip, May and Cadwgan, Jill},
  journal={Paediatrics and Child Health},
  volume={34},
  number={8},
  pages={279--289},
  year={2024},
  publisher={Elsevier}
}

@article{makki2014prevalence,
  title={Prevalence and pattern of upper limb involvement in cerebral palsy},
  author={Makki, Daoud and Duodu, J and Nixon, Matthew},
  journal={Journal of children's orthopaedics},
  volume={8},
  number={3},
  pages={215--219},
  year={2014},
  publisher={SAGE Publications Sage UK: London, England}
}

@article{de2016kinematic,
  title={Kinematic upper limb evaluation of children and adolescents with cerebral palsy: a systematic review of the literature},
  author={de Moura, Renata Calhes Franco and Almeida, Cibele Santos and Dumont, Arislander Jonatan Lopes and Lazzari, Roberta Delasta and Lopes, Jamile Benite Palma and de Carvalho Duarte, Natalia Almeida and Braun, Luiz Ferreira and Oliveira, Claudia Santos},
  journal={Journal of physical therapy science},
  volume={28},
  number={2},
  pages={695--700},
  year={2016},
  publisher={The Society of Physical Therapy Science}
}

@article{chen2014effectiveness,
  title={Effectiveness of constraint-induced movement therapy on upper-extremity function in children with cerebral palsy: a systematic review and meta-analysis of randomized controlled trials},
  author={Chen, Yu-ping and Pope, Stephanie and Tyler, Dana and Warren, Gordon L},
  journal={Clinical Rehabilitation},
  volume={28},
  number={10},
  pages={939--953},
  year={2014},
  publisher={SAGE Publications Sage UK: London, England}
}

@article{novak2013systematic,
  title={A systematic review of interventions for children with cerebral palsy: state of the evidence},
  author={Novak, Iona and Mcintyre, Sarah and Morgan, Catherine and Campbell, Lanie and Dark, Leigha and Morton, Natalie and Stumbles, Elise and Wilson, Salli-Ann and Goldsmith, Shona},
  journal={Developmental medicine \& child neurology},
  volume={55},
  number={10},
  pages={885--910},
  year={2013},
  publisher={Wiley Online Library}
}

@article{lopes2018games,
  title={Games used with serious purposes: a systematic review of interventions in patients with cerebral palsy},
  author={Lopes, S{\'\i}lvia and Magalh{\~a}es, Paula and Pereira, Armanda and Martins, Juliana and Magalh{\~a}es, Carla and Chaleta, Elisa and Ros{\'a}rio, Pedro},
  journal={Frontiers in psychology},
  volume={9},
  pages={1712},
  year={2018},
  publisher={Frontiers Media SA}
}

@article{ahn2023scoping,
  title={A scoping review of the serious game-based rehabilitation of people with cerebral palsy},
  author={Ahn, Si Nae},
  journal={International Journal of Environmental Research and Public Health},
  volume={20},
  number={21},
  pages={7006},
  year={2023},
  publisher={MDPI}
}

@inproceedings{garcia2013super,
  title={Super pop vrtm: An adaptable virtual reality game for upper-body rehabilitation},
  author={Garc{\'\i}a-Vergara, Sergio and Chen, Yu-Ping and Howard, Ayanna M},
  booktitle={international conference on virtual, augmented and mixed reality},
  pages={40--49},
  year={2013},
  organization={Springer}
}

@article{chen2018effectiveness,
  title={Effectiveness of virtual reality in children with cerebral palsy: a systematic review and meta-analysis of randomized controlled trials},
  author={Chen, Yuping and Fanchiang, HsinChen D and Howard, Ayanna},
  journal={Physical therapy},
  volume={98},
  number={1},
  pages={63--77},
  year={2018},
  publisher={Oxford University Press}
}

@article{demers2021integration,
  title={Integration of motor learning principles into virtual reality interventions for individuals with cerebral palsy: systematic review},
  author={Demers, Marika and Fung, Karen and Subramanian, Sandeep K and Lemay, Martin and Robert, Maxime T},
  journal={JMIR Serious Games},
  volume={9},
  number={2},
  pages={e23822},
  year={2021},
  publisher={JMIR Publications Toronto, Canada}
}

@article{chen2016effects,
  title={Effects of robotic therapy on upper-extremity function in children with cerebral palsy: a systematic review},
  author={Chen, Yu-Ping and Howard, Ayanna M},
  journal={Developmental neurorehabilitation},
  volume={19},
  number={1},
  pages={64--71},
  year={2016},
  publisher={Taylor \& Francis}
}

@inproceedings{jeglinsky2024rehabilitation,
  title={Rehabilitation with Humanoid Robots: A Feasibility Study of Rehabilitation of Children with Cerebral Palsy (CP) Using a QTRobot},
  author={Jeglinsky-Kankainen, Ira and Hellst{\'e}n, Thomas and Karlsson, Jonny and Espinosa-Leal, Leonardo},
  booktitle={International Conference on Smart Technologies \& Education},
  pages={390--400},
  year={2024},
  organization={Springer}
}

@article{van2017robot,
  title={Robot ZORA in rehabilitation and special education for children with severe physical disabilities: a pilot study},
  author={van den Heuvel, Ren{\'e}e JF and Lexis, Monique AS and de Witte, Luc P},
  journal={International journal of rehabilitation research. Internationale Zeitschrift fur Rehabilitationsforschung. Revue internationale de recherches de readaptation},
  volume={40},
  number={4},
  pages={353},
  year={2017}
}

@inproceedings{kozyavkin2014humanoid,
  title={Humanoid social robots in the rehabilitation of children with cerebral palsy.},
  author={Kozyavkin, Volodymyr and Kachmar, Oleh and Ablikova, Iryna},
  booktitle={PervasiveHealth},
  pages={430--431},
  year={2014}
}

@article{chen2018effect,
  title={Effect of feedback from a socially interactive humanoid robot on reaching kinematics in children with and without cerebral palsy: a pilot study},
  author={Chen, Yuping and Garcia-Vergara, Sergio and Howard, Ayanna M},
  journal={Developmental neurorehabilitation},
  volume={21},
  number={8},
  pages={490--496},
  year={2018},
  publisher={Taylor \& Francis}
}

@inproceedings{lee2017does,
  title={Does appearance matter? Validating engagement in therapy protocols with socially interactive humanoid robots},
  author={Lee, Breanna and Xu, Jin and Howard, Ayanna},
  booktitle={2017 IEEE symposium series on computational intelligence (SSCI)},
  pages={1--6},
  year={2017},
  organization={IEEE}
}

@inproceedings{xu2018investigating,
  title={Investigating the relationship between believability and presence during a collaborative cognitive task with a socially interactive robot},
  author={Xu, Jin and Howard, Ayanna},
  booktitle={2018 27th IEEE international symposium on robot and human interactive communication (RO-MAN)},
  pages={137--143},
  year={2018},
  organization={IEEE}
}

@inproceedings{xu2018robot,
  title={Robot therapist versus human therapist: Evaluating the effect of corrective feedback on human motor performance},
  author={Xu, Jin and De'Aira, G Bryant and Chen, Yu-Ping and Howard, Ayanna},
  booktitle={2018 International Symposium on Medical Robotics (ISMR)},
  pages={1--6},
  year={2018},
  organization={IEEE}
}

@inproceedings{de2019effect,
  title={The effect of robot vs. human corrective feedback on children's intrinsic motivation},
  author={De'Aira, G Bryant and Xu, Jin and Chen, Yu-Ping and Howard, Ayanna},
  booktitle={2019 14th ACM/IEEE International Conference on Human-Robot Interaction (HRI)},
  pages={638--639},
  year={2019},
  organization={IEEE}
}

@article{openai2023gpt4,
  title   = {GPT-4 Technical Report},
  author  = {{OpenAI} and Achiam, Josh and Adler, Steven and Agarwal, Sandhini and Ahmad, Lama and Akkaya, Ilge and Aleman, Florencia Leoni and Almeida, Diogo and Altenschmidt, Janko and Altman, Sam and others},
  journal = {arXiv preprint arXiv:2303.08774},
  year    = {2023},
  url     = {https://arxiv.org/abs/2303.08774}
}

@misc{microsoft_azure_kinect_dk,
  author       = {{Microsoft Azure Kinect DK}},
  title        = {Azure Kinect DK},
  howpublished = {\url{https://azure.microsoft.com/en-us/products/kinect-dk}},
}

@misc{orbbec_femto_bolt,
  author       = {{Orbbec Femto Bolt}},
  title        = {Femto Bolt},
  howpublished = {\url{https://www.orbbec.com/products/tof-camera/femto-bolt/}},
}

@misc{hello_robotics,
  author       = {{Hello Robotics}},
  title        = {Hello Robotics},
  howpublished = {\url{https://www.hello-robo.com/}},
}
\bibliographystyle{plainnat}

\newpage
\appendix
\section{Additional material}
\label{app:extra}

\begin{longtable}{p{0.9\linewidth}}
\caption{GPT-4-generated utterance variations: Introduction command. The first row is the original utterance from the SuperPop-NAO platform; the remaining ten rows are GPT-4-generated variations preserving the same meaning.} \label{tab:utterances_intro} \\
\toprule
\endfirsthead

\toprule
\textbf{Utterance} \\
\midrule
\endhead

\bottomrule
\endfoot

\bottomrule
\endlastfoot

\textbf{[Original]} Hello. My name is Thrive and I will be playing Super Pop with you today. I will ask you to complete a series of tasks and I would love it if you could follow my instructions. When you are ready, please raise both of your hands as high as you can. \\
\midrule
Hi there! I'm Thrive, and today we're going to play Super Pop! I'll guide you through some fun tasks, and I'd love for you to follow my instructions. When you're ready, raise both hands as high as you can! \\
Hello! My name is Thrive, and I'm super excited to play Super Pop with you today! I have some fun tasks for you, so when you're ready, lift both hands up as high as possible! \\
Hey there! I'm Thrive, your game buddy for today! We're going to play Super Pop, and I'll give you some fun challenges. Ready? Raise both hands up high! \\
Hi! I'm Thrive, and I can't wait to play Super Pop with you! I'll be giving you some fun tasks—when you're ready, reach up high with both hands! \\
Hello, my friend! I'm Thrive, and today we'll play Super Pop together! I'll guide you through some awesome tasks. If you're ready, stretch both hands up high! \\
Hey! I'm Thrive, and I'm thrilled to play Super Pop with you today! I have some cool challenges for you—when you're ready, raise both hands as high as you can! \\
Hi! I'm Thrive, and I'll be your game buddy today! Super Pop is going to be so much fun! Let's start—raise your hands high when you're ready! \\
Hello! My name is Thrive, and I'm excited to play Super Pop with you! I'll be giving you instructions—when you're ready, lift your hands way up high! \\
Hey, superstar! I'm Thrive, and we're going to have a blast playing Super Pop! I'll guide you through some tasks, so when you're ready, stretch those hands up high! \\
Hi there! I'm Thrive! Today, we're going to play Super Pop together! Follow my instructions and, when you're ready, raise both hands as high as you can! \\

\end{longtable}

\begin{longtable}{p{0.9\linewidth}}
\caption{GPT-4-generated utterance variations: ``Move faster'' command. The first row is the original utterance from the SuperPop-NAO platform; the remaining rows are GPT-4-generated variations preserving the same meaning.} \label{tab:utterances_faster} \\
\toprule
\endfirsthead

\toprule
\endhead

\bottomrule
\endfoot

\bottomrule
\endlastfoot

\textbf{[Original]} Keep up the good work. Move a little faster. \\
\midrule
You're doing great! Let's pick up the pace a little! \\
Awesome job! Try moving just a bit faster! \\
Keep going! Can you speed up a little? \\
You're on fire! Let's go a tiny bit quicker! \\
Wow, you're doing amazing! Try moving a bit faster! \\
Fantastic effort! Let's see if you can go a little quicker! \\
Great work! Can you move just a little bit faster? \\
You're rocking this! Let's speed up a tiny bit! \\
Keep it up! Try picking up the pace a little! \\
You're doing so well! Let's see if you can move a bit faster! \\
Amazing energy! Can we speed things up a tiny bit? \\
You're crushing it! Try going just a little faster! \\
Fantastic! Let's see if you can move a little quicker! \\
Keep up the good work! Can you go a bit faster? \\
You're doing great! Let's try a slightly faster pace! \\
Impressive! See if you can speed up just a bit! \\
Nice job! Let's go a little faster this time! \\
You're a superstar! Try moving just a bit quicker! \\
Wonderful effort! Let's add a little more speed! \\
Great going! Let's push the pace a tiny bit more! \\
You're doing fantastic! Let's pick up the speed just a little! \\
Great job! Can you go just a tiny bit faster? \\
Super work! Let's see if we can move a little quicker! \\
You're making great progress! Try speeding up a bit! \\
Keep it up! Let's add a little more energy and go faster! \\
You're unstoppable! See if you can move just a bit quicker! \\
Amazing effort! Let's push the pace just a little more! \\
Way to go! Can we pick up the speed just a tiny bit? \\
Keep that momentum going! Let's move a little faster! \\
You're doing so well! Let's speed things up just a little! \\

\end{longtable}

\begin{longtable}{p{0.9\linewidth}}
\caption{GPT-4-generated utterance variations: ``Keep the same speed'' command. The first row is the original utterance from the SuperPop-NAO platform; the remaining rows are GPT-4-generated variations preserving the same meaning.} \label{tab:utterances_slower} \\
\toprule
\endfirsthead

\toprule
\endhead

\bottomrule
\endfoot

\bottomrule
\endlastfoot

\textbf{[Original]} Fantastic! Let us move at the exact same speed. \\
\midrule
Fantastic! Keep moving at this exact speed! \\
Great job! Maintain this pace! \\
Awesome! Stay at this speed, just like that! \\
You're doing great! Keep your movement steady! \\
Perfect! Let's continue at this pace! \\
Nice work! Keep going at this speed! \\
Super! Hold this rhythm steady! \\
Wonderful! Maintain this exact speed! \\
You're crushing it! Keep up this steady pace! \\
Excellent! Stay at this same speed! \\
Amazing work! Keep moving just like this! \\
You're doing fantastic! No need to speed up or slow down! \\
Keep going! Stay right at this pace! \\
That's it! Keep your movement smooth and steady! \\
Great effort! Let's keep this exact rhythm! \\
You're nailing it! Stick with this speed! \\
Stay strong! Keep this pace steady! \\
Well done! Keep your movement just like this! \\
Awesome work! Let's keep this same speed! \\
You've got it! Maintain this steady pace! \\
Perfect speed! Just keep moving like this! \\
You're a superstar! Keep the same rhythm! \\
Brilliant! Keep up this perfect pace! \\
Just right! Stay steady at this speed! \\
You're doing so well! Keep your movements at this tempo! \\
Keep rocking! No need to change your speed! \\
Way to go! Keep your pace just like this! \\
You're on fire! Maintain this awesome speed! \\
You've got the rhythm! Keep it just like this! \\
Steady and strong! Keep moving just like that! \\

\end{longtable}

\begin{longtable}{p{0.9\linewidth}}
\caption{GPT-4-generated utterance variations: Game-end feedback command. The first row is the original utterance from the SuperPop-NAO platform; the remaining rows are GPT-4-generated variations preserving the same meaning.} \label{tab:utterances_end} \\
\toprule
\endfirsthead

\toprule
\textbf{Utterance} \\
\midrule
\endhead

\bottomrule
\endfoot

\bottomrule
\endlastfoot

\textbf{[Original]} Fantastic! Wow! Good game. Let us play another one. \\
\midrule
Fantastic! Wow! That was a great game! Let's play another one! \\
Awesome job! That was so much fun! Ready for another round? \\
You did amazing! What a great game! Let's go again! \\
Wow! That was fantastic! How about another game? \\
Great work! You rocked that game! Let's play again! \\
Super effort! That was fun! Want to play one more? \\
Amazing! You're so good at this! Let's do another game! \\
That was incredible! Let's go for another round! \\
Woohoo! You did awesome! Let's play again! \\
Nice job! What a fun game! How about another one? \\

\end{longtable}


\end{document}